\documentclass[final,a4paper,oneside]{paper}

\usepackage[final]{graphicx}
\usepackage[left=20mm,right=20mm,top=30mm,bottom=30mm]{geometry}
\usepackage[normal,small,bf]{caption2}
\usepackage{algorithmic}
\usepackage[section]{algorithm}
\usepackage{cite}

\newtheorem{definition}{Definition}[section]

\begin{document}

\title{Randomised Variable Neighbourhood Search for Multi Objective
Optimisation}
\author{Martin Josef Geiger}
\institution{School of Computer Science \& IT, University of
Nottingham, mjg@cs.nott.ac.uk}
\date{}
\maketitle

\begin{abstract}
Various local search approaches have recently been applied to
machine scheduling problems under multiple objectives. Their
foremost consideration is the identification of the
set of Pareto optimal alternatives. An important aspect of
successfully solving these problems lies in the definition of an
appropriate neighbourhood structure. Unclear in this context
remains, how interdependencies within the fitness landscape affect
the resolution of the problem.

The paper presents a study of neighbourhood search operators for
multiple objective flow shop scheduling. Experiments have been
carried out with twelve different combinations of criteria. To
derive exact conclusions, small problem instances, for which the
optimal solutions are known, have been chosen. Statistical tests
show that no single neighbourhood operator is able to equally
identify all Pareto optimal alternatives. Significant improvements
however have been obtained by hybridising the solution algorithm
using a randomised variable neighbourhood search technique.
\end{abstract}


\section{Introduction}
Machine scheduling considers in general the assignment of a set of
resources (machines) $\mathcal{M} = \{M_{1}, \ldots, M_{m} \}$ to
a set of jobs $\mathcal{J} = \{ J_{1}, \ldots, J_{n} \}$, each of
which consists of a set of operations $J_{j} = \{ O_{j1}, \ldots,
O_{jo_{j}} \}$ \cite{blazewicz:2001:book}. The operations $O_{jk}$
typically may be processed on a single machine $M_{i} \in
\mathcal{M} $ involving a nonnegative processing time $t_{jk}$.
Usually, precedence constraints are defined among the operations
of a job, reflecting its technical nature. In the specific case of
the here considered \emph{permutation flow shop scheduling
problem}, the machine sequences are identical for all jobs
\cite{pinedo:2002:book}. Furthermore, the sequence of the jobs on
the machines is assumed to be the same, therefore leading to a
single job permutation as a possible schedule representation. A
solution of the problem, a \emph{schedule}, defines starting times
for the tasks respecting the chosen job sequence and the defined
constraints of the problem.

Optimality of schedules can be judged with respect to a single or
multiple objective functions, expressing the quality of the
solution in a quantitative way. While the multi criteria nature of
scheduling in manufacturing environments has been mentioned quite
early \cite{rinnooykan:1976:book}, the problem is treated only
recently as a multi objective optimisation problem
\cite{tkindt:2002:book}. Here, the goal is to identify all
efficient alternatives, the \emph{set of Pareto optimal solutions
$P$} \cite{vanveldhuizen:2000:article}. Most important optimality
criteria are based on the completion times $C_{j}$ of the jobs
$J_{j}$ in the schedule:
\begin{itemize}
    \item The minimisation of the maximum completion time (makespan)
    $C_{max} = \max \{ C_{1}, \ldots, C_{n} \} $.
    \item The minimisation of the sum of the completion times $C_{sum} =
    \sum_{j=1}^{n} C_{j} $
\end{itemize}
In the case of existing due dates $d_{j}$ for each job $J_{j}$, it
is possible to compute due date violations in the form of
tardiness values $T_{j} = \max \{ C_{j} - d_{j}, 0 \}$. Possible
optimality criteria are:
\begin{itemize}
    \item The minimisation of the maximum tardiness $T_{max} = \max \{ T_{1}, \ldots,
    T_{j} \}$.
    \item The minimisation of the total tardiness $T_{sum} = \sum_{j=1}^{n}
    T_{j}$.
    \item The minimisation of the number of tardy jobs $U = \sum_{j=1}^{n}
    U_{j}$ where $U_{j} = \left\{ \begin{array} {r@{\quad:\quad}l} 1 & T_{j} > 0 \\ 0 & T_{j} = 0 \end{array}
    \right.$
\end{itemize}
In terms of machine efficiency, idle times $I_{i}$ of the machines
$M_{i}$  may be considered up to the completion of the last job.
Optimality criteria are therefore:
\begin{itemize}
    \item The minimisation of the maximum machine idleness $I_{max} = \max \{I_{1},
    \ldots, I_{m} \}$.
    \item The minimisation of the total machine idleness $I_{sum} =
    \sum_{i=1}^{m} I_{i}$.
\end{itemize}

An important factor for the resolution of the permutation flow
shop scheduling problem is the circumstance that the functions are
\emph{regular} \cite{conway:1967:book}. It is therefore possible
to map a given sequence of jobs $\pi = \{ \pi_{1}, \ldots, \pi_{n}
\} $ into an active schedule, one of which is optimal
\cite{daniels:1993:article}. Consequently, many existing local
search approaches are based on this idea of a schedule
representation, and neighbourhood definitions manipulate the job
permutations $\pi$ by changing the positions of jobs
\cite{reeves:1999:article}:
\begin{itemize}
    \item Exchange neighbourhood (EX), exchanging the position of
    two jobs $\pi_{j}$ and $\pi_{k}$, $j \neq k$.
    \item Forward shift neighbourhood (FSH), removing a job $\pi_{j}$
    and reinserting it at position $\pi_{k}$ with $k > j$.
    \item Backward shift neighbourhood (BSH), removing a job
    $\pi_{j}$ and reinserting it at position $\pi_{k}$ with $k < j$.
    \item Inversion neighbourhood (INV), inverting the positions of
    a chosen subset of $\pi$.
\end{itemize}

Important applications  of local search algorithms for the
permutation flow shop scheduling problem comprise local search
descent
\cite{werner:1993:article,glass:1996:article,reeves:1999:article},
simulated annealing
\cite{osman:1989:article,glass:1996:article,sotskov:1996:article,ogbu:1990:article,ishibuchi:1995:article,gangadharan:1994:article},
tabu search
\cite{widmer:1989:article,taillard:1990:article,reeves:1993:article},
and evolutionary algorithms
\cite{bierwirth:1993:book,nagar:1996:article,stoppler:1992:incollection,murata:1996:article,basseur:2002:inproceedings,ishibuchi:1998:article,ishibuchi:2003:article,bagchi:1999:book,bagchi:2001:inproceedings}.
The majority of the previously conducted studies consider the
problem as a single objective optimisation problem with respect to
the makespan objective $C_{max}$. Here, shift neighbourhood
operators lead in most cases to superior results
\cite{taillard:1990:article}. A generalisation of the conclusions
with respect to multi objective problem settings seems however
questionable, and corresponding studies have not been conducted
yet.

\section{\label{sec:investigation:local:search}An investigation of neighbourhood operators}
\subsection{Local search framework and experimental setup}
The effectiveness of local search operators for multi objective
flow shop scheduling problems has been investigated using the
framework described in algorithm \ref{alg:MOLSD}
\cite{talbi:2001:inproceedings}.

\begin{algorithm}[ht]
\caption{\label{alg:MOLSD}Multi Objective Local Search Descent}
\begin{algorithmic}[1]%
\STATE Generate initial solution $x$%
\STATE $P^{approx} = \{ x \}$%
\REPEAT%
    \STATE Select $x \in P^{approx}$ for which $nh(x)$ has not been investigated yet%
    \STATE Generate $nh(x)$%
    \STATE Update $P^{approx}$ with all $x' \in nh(x) $%
    \IF{$x \in P^{approx}$}%
        \STATE Mark $nh(x)$ as \rq{}investigated\lq{}%
    \ENDIF%
\UNTIL{$\not\!\exists x \in P^{approx}$ with $nh(x)$ still to be investigated}%
\end{algorithmic}
\end{algorithm}

Starting from a random initial solution, neighbouring solutions
are generated until no further improvement is possible. With
respect to the goal of approximating a Pareto set $P$ of
alternatives, an approximation set $P^{approx}$ is maintained
during search. The update of $P^{approx}$ is done according to the
the Pareto dominance relation, given in definition
\ref{def:dominance} for the chosen objective vector $G(x) =
(g_{1}(x), \ldots, g_{k}(x))$. $P^{approx}$ therefore only
contains nondominated alternatives. Solutions $x' \in nh(x)$ that
dominate $x$ lead to the removal of $x$ from $P^{approx}$, making
step 7 after the update of $P^{approx}$ necessary.

\begin{definition}[Pareto dominance]\label{def:dominance}
A vector of objective functions $G(x)$ is said to dominate an
vector $G(x')$ if and only if $\forall i\,\,\, g_{i}(x)\leq
g_{i}(x') \wedge \exists i \mid g_{i}(x) < g_{i}(x')$.
\end{definition}

The neighbourhood operators applied within the framework have been
$k$-FSH, $k$-BSH, and $k$-EX. A control parameter $k$ determines
the number of succeeding jobs in the sequence being involved in
the shift or exchange operation, resulting in a block shift or
block exchange neighbourhood for $k > 1$. Settings of $k = 1$, $k
= 2$, and $k = 3$ have been tested. 100 Test instances have been
generated with $n = m = 10$ following the proposal of {\sc
Taillard} \cite{taillard:1993:article}, and the due dates of the
jobs were generated adopting the methodology of {\sc Demirkol et
al.} \cite{demirkol:1998:article}. The size of the instances
allows the determination of the true Pareto set by enumerating all
possible alternatives within reasonable time.

\begin{table}[ht]\centering
\caption[Übersicht der getesteten
Problemklassen]{\label{tbl:getestete:Problemklassen}Investigated
combinations of optimality criteria.}
\begin{tabular}{ccl}\\
\hline%
Abbreviation & No objectives & Classification $\alpha \mid \beta \mid \gamma$ \cite{graham:1979:article}\\
\hline
$\gamma_{1}$ & 2 & $F \mid prmu, d_{j} \mid C_{max}, T_{max}$\\
$\gamma_{2}$ & 2 & $F \mid prmu, d_{j} \mid C_{max}, C_{sum}$\\
$\gamma_{3}$ & 2 & $F \mid prmu, d_{j} \mid C_{max}, T_{sum}$\\
$\gamma_{4}$ & 2 & $F \mid prmu, d_{j} \mid T_{max}, T_{sum}$\\
$\gamma_{5}$ & 2 & $F \mid prmu, d_{j} \mid C_{sum}, T_{max}$\\
$\gamma_{6}$ & 2 & $F \mid prmu, d_{j} \mid C_{sum}, T_{sum}$\\
$\gamma_{7}$ & 3 & $F \mid prmu, d_{j} \mid C_{max}, T_{max}, T_{sum}$\\
$\gamma_{8}$ & 3 & $F \mid prmu, d_{j} \mid C_{max}, C_{sum}, T_{max}$\\
$\gamma_{9}$ & 3 & $F \mid prmu, d_{j} \mid C_{max}, C_{sum}, T_{sum}$\\
$\gamma_{10}$ & 3 & $F \mid prmu, d_{j} \mid C_{sum}, T_{max}, T_{sum}$\\
$\gamma_{11}$ & 4 & $F \mid prmu, d_{j} \mid C_{max}, C_{sum}, T_{max}, T_{sum}$\\
$\gamma_{12}$ & 6 & $F \mid prmu, d_{j} \mid C_{max}, C_{sum}, T_{max}, T_{sum}, I_{sum}, U$\\
\hline
\end{tabular}
\end{table}

Each neighbourhood operator has been tested on each problem
instance in 100 test runs regarding twelve different combinations
of optimality criteria as given in table
\ref{tbl:getestete:Problemklassen}, leading to a total of
1,080,000 local search runs. The approximation quality of the
obtained set $P^{approx}$ to the Pareto set $P$ has been measured
using the $D_{1}$ (average deviation of $P^{approx}$ to $P$) and
$D_{2}$ (maximum deviation of $P^{approx}$ to $P$) metrics of {\sc
Czy\.{z}ak} and {\sc Jaszkiewicz} \cite{czyzak:1998:article}.

\subsection{Results}
The average results obtained by the local search operators have
been investigated for statistical significance using a t-test at a
given level of significance of 0.01. Table
\ref{tbl:Ergebnisse:MOLSD:Daver} and
\ref{tbl:Ergebnisse:MOLSD:Dmax} give the number of test instances,
in which a neighbourhood definition led to a significantly best
approximation for $D_{1}$ and $D_{2}$.

\begin{table}[ht]\centering
\caption{\label{tbl:Ergebnisse:MOLSD:Daver}Significance test
results for $D_{1}$
over all 100 test instances.} %

\begin{tabular}{lrrrrrrrrr}\\%
\hline%
$\gamma$ & 1-BSH & 2-BSH & 3-BSH & 1-FSH & 2-FSH &
3-FSH & 1-EX & 2-EX & 3-EX\\
\hline
$\gamma_{1}$ & 0 & 0 & 0  & 37  & 0  & 0  & 15  & 0  & 0 \\
$\gamma_{2}$ & 20 & 0 & 0  & 9  & 0  & 0  & 10 & 0  & 0 \\
$\gamma_{3}$ & 11 & 0 & 0  & 9  & 0  & 0  & 16  & 0  & 0 \\
$\gamma_{4}$ & 2 & 0 & 0  & 30  & 0  & 0  & 13  & 0  & 0 \\
$\gamma_{5}$ & 1 & 0  & 0  & 39  & 0  & 0  & 15  & 0  & 0 \\
$\gamma_{6}$ & 32 & 0  & 0  & 1  & 0  & 0  & 15  & 0  & 0 \\
$\gamma_{7}$ & 2 & 0  & 0  & 44  & 0  & 0  & 19  & 0  & 0 \\
$\gamma_{8}$ & 3 & 0  & 0  & 48  & 0  & 0  & 18  & 0  & 0 \\
$\gamma_{9}$ & 20  & 0  & 0  & 17  & 0  & 0  & 17  & 0  & 0 \\
$\gamma_{10}$ & 6  & 0  & 0  & 29  & 0  & 0  & 15  & 0  & 0 \\
$\gamma_{11}$ & 8  & 0  & 0  & 47  & 0  & 0  & 18  & 0  & 0 \\
$\gamma_{12}$ & 5 & 0  & 0  & 42  & 0  & 0  & 37  & 0  & 0 \\
\hline%
\end{tabular}
\end{table}

\begin{table}[ht]\centering
\caption{\label{tbl:Ergebnisse:MOLSD:Dmax}Significance test
results for $D_{2}$ over all 100 test instances.}
\begin{tabular}{lrrrrrrrrr}\\%
\hline%
$\gamma$ & 1-BSH & 2-BSH & 3-BSH & 1-FSH & 2-FSH &
3-FSH & 1-EX & 2-EX & 3-EX\\
\hline
$\gamma_{1}$ & 0  & 0  & 0  & 53  & 1  & 0  & 23  & 0  & 0\\
$\gamma_{2}$ & 22  & 0  & 0  & 14  & 0  & 0  & 12  & 0  & 0\\
$\gamma_{3}$ & 13  & 0  & 0  & 15  & 0  & 0  & 24  & 0  & 0\\
$\gamma_{4}$ & 3  & 0  & 0  & 41  & 0  & 0  & 20  & 0  & 0 \\
$\gamma_{5}$ & 4  & 0  & 0  & 49  & 1  & 0  & 24  & 0 & 0 \\
$\gamma_{6}$ & 30  & 0  & 0  & 2  & 0  & 0  & 19  & 0  & 0 \\
$\gamma_{7}$ & 2  & 0  & 0  & 49  & 2  & 0  & 24  & 0  & 0 \\
$\gamma_{8}$ & 5  & 1  & 0  & 56  & 0  & 0  & 19  & 0  & 0 \\
$\gamma_{9}$ & 20  & 0  & 0  & 23  & 0  & 0  & 23  & 0  & 0 \\
$\gamma_{10}$ & 14  & 1  & 0  & 34  & 0  & 0  & 27  & 0  & 0 \\
$\gamma_{11}$ & 10  & 2  & 0  & 51  & 1  & 0  & 21  & 0  & 0 \\
$\gamma_{12}$ & 10  & 1  & 0  & 32  & 0  & 0  & 53  & 0  & 0 \\
\hline%
\end{tabular}
\end{table}

It is possible to observe that neighbourhoods with a control
parameter of $k > 1$ are only superior for very few problem
instances. Comparing 1-FSH, 1-BSH and 1-EX, no operator turns out
to be the most appropriate in all cases. While forward shift
tends to lead more frequently to significantly best results, there
are still numerous instances for which other operators are more
favourable. It can also be noticed that this result is not
primarily depending on the optimality criteria involved.

\begin{figure}[ht]
\centerline{
\includegraphics[width=6.5cm]{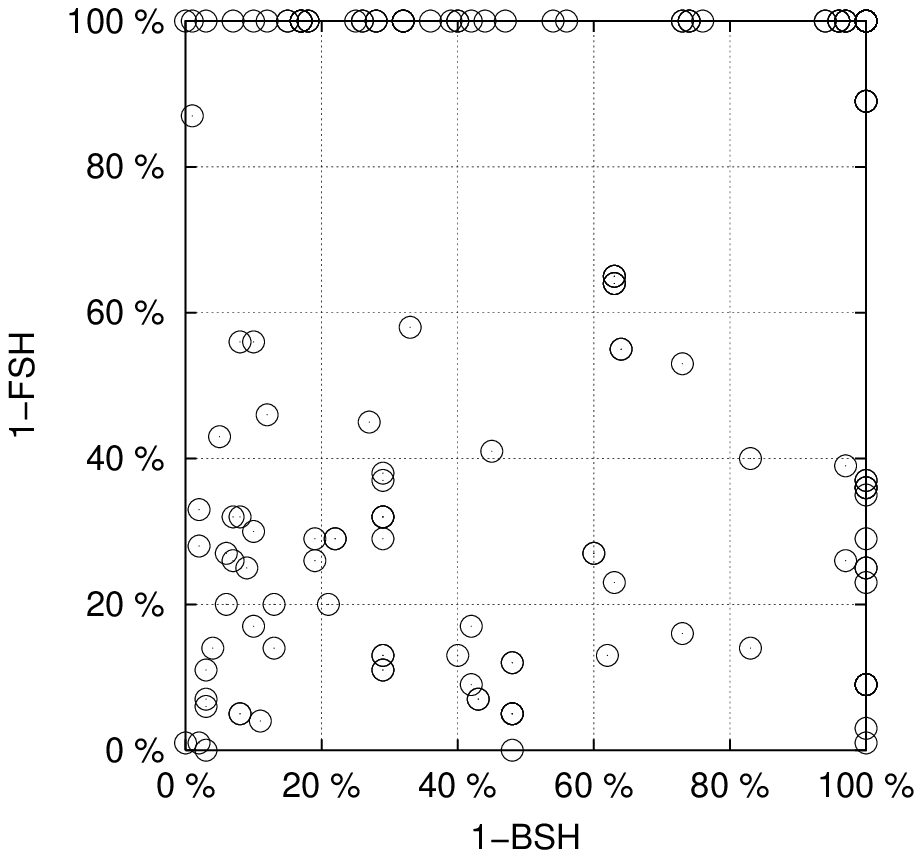}
\includegraphics[width=6.5cm]{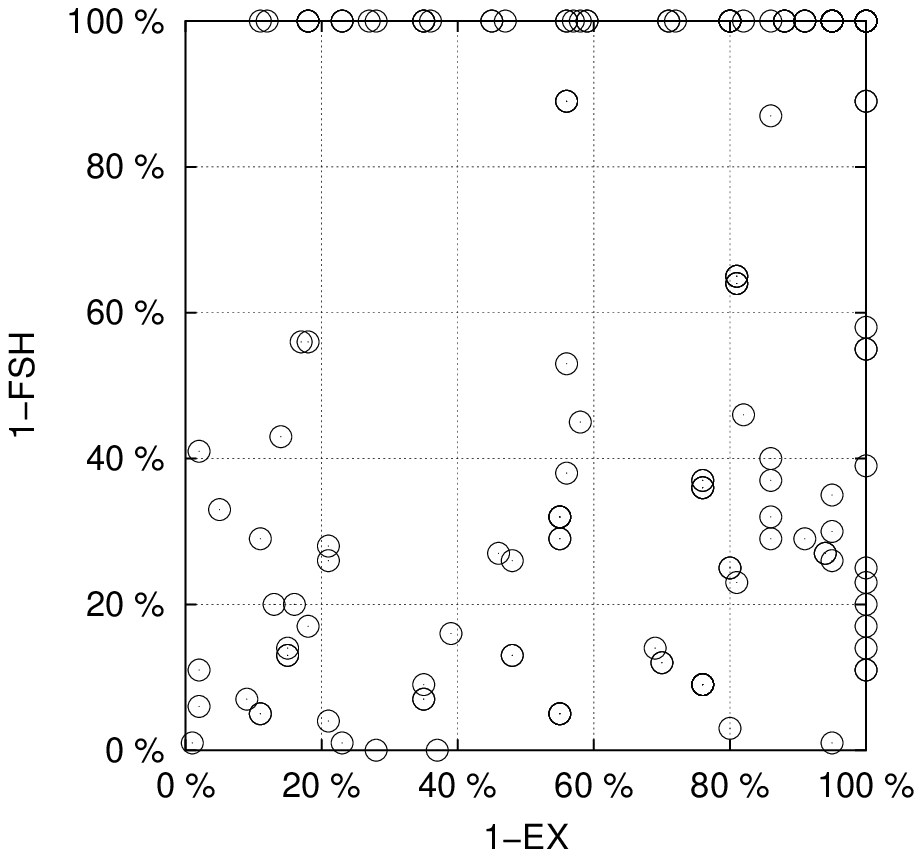}}
\caption{\label{fig:Haeufigkeit:Gefunden}Frequency of
identification of Pareto optimal alternatives depending on the
neighbourhood operator.}
\end{figure}

On a more detailed level it is possible to observe that the
identification probability of a single Pareto optimal alternative
depends on the choice of the neighbourhood operator. To illustrate
this circumstance, figure \ref{fig:Haeufigkeit:Gefunden} plots for
each $x \in P$ of a problem instance, involving the optimality
criteria combination $\gamma_{11}$, the frequency of its
identification. No correlation between the identification
frequencies can be seen, and the analysis reveals the existence of
Pareto optimal alternatives which may be identified rather easily
using e.g. 1-FSH while not being found by the 1-EX neighbourhood
and vice versa.

\section{Multi Objective Variable Neighbourhood Search}
\subsection{Description of the approach}
Based on the investigation presented in section
\ref{sec:investigation:local:search}, an improved local search
approach has been proposed, described in algorithm
\ref{alg:MOVNS}. Its main idea is the randomised application of a
set of neighbourhood operators to increase the overall
identification probability of every Pareto optimal alternative.
Compared with existing concepts of variable neighbourhood search
\cite{mladenovic:1997:article}, the applied operator is randomly
chosen at every step of the algorithm from a predefined set of
neighbourhoods. Also, only a single neighbourhood is generated,
even if it is not possible to achieve further improvements with
the selected operator. The computational complexity of this
approach is therefore identical with the multi objective local
search framework in section \ref{sec:investigation:local:search}.

\begin{algorithm}[ht]
\caption{\label{alg:MOVNS}Multi Objective Variable Neighbourhood
Search (MOVNS)}
\begin{algorithmic}[1]%
\STATE Generate initial solution $x$%
\STATE $P^{approx} = \{ x \}$%
\REPEAT%
    \STATE Select $x \in P^{approx}$ for which $nh(x)$ has not been investigated yet%
    \STATE Select a neighbourhood $nh$%
    \STATE Generate $nh(x)$%
    \STATE Update $P^{approx}$ with all $x' \in nh(x) $%
    \IF{$x \in P^{approx}$}%
        \STATE Mark $nh(x)$ as \rq{}investigated\lq{}%
    \ENDIF%
\UNTIL{$\not\!\exists x \in P^{approx}$ with $nh(x)$ still to be investigated}%
\end{algorithmic}
\end{algorithm}

The MOVNS heuristic has been applied to the 100 test instances
regarding the twelve optimality criteria definitions in 100 test
runs each. Two different configurations have been considered:
\begin{enumerate}
    \item MOVNS/3, applying a neighbourhood from the set \{1-BSH,
    1-FSH, 1-EX\}.
    \item MOVNS/9, selecting in each step from all nine
    neighbourhood definitions as given in section
    \ref{sec:investigation:local:search}.
\end{enumerate}

\subsection{Results}
Statistical tests of significance show that MOVNS/3 is able to
outperform the multi objective local search approach involving a
single operator in most test instances. Table
\ref{tbl:Ergebnisse:MOLSD:Daver:VNS} and
\ref{tbl:Ergebnisse:MOLSD:Dmax:VNS} give the number of problem
instances in which the approach led to significantly best results
for $D_{1}$ and $D_{2}$. This behaviour does not seem to depend on
the defined optimality criteria. For a few problem instances
however, and especially for the maximum deviation $D_{2}$, no
improvements are possible.

\begin{table}[ht]\centering
\caption{\label{tbl:Ergebnisse:MOLSD:Daver:VNS}Significance test
results for $D_{1}$ over all 100 test instances.}
\begin{tabular}{lrrrrr}\\%
\hline%
$\gamma$      & 1-BSH & 1-FSH & 1-EX & MOVNS/3        & MOVNS/9\\
\hline
$\gamma_{1}$  & 0   & 1   & 1   & 53  & 0 \\
$\gamma_{2}$  & 3   & 3   & 0   & 47  & 0 \\
$\gamma_{3}$  & 0   & 2   & 0   & 53  & 0 \\
$\gamma_{4}$  & 0   & 4   & 0   & 35  & 0 \\
$\gamma_{5}$  & 0   & 8   & 0   & 46  & 0 \\
$\gamma_{6}$  & 3   & 0   & 1   & 30  & 0 \\
$\gamma_{7}$  & 0   & 7   & 0   & 63  & 0 \\
$\gamma_{8}$  & 0   & 8   & 0   & 65  & 0 \\
$\gamma_{9}$  & 0   & 8   & 0   & 65  & 0 \\
$\gamma_{10}$ & 1   & 6   & 0   & 54  & 0 \\
$\gamma_{11}$ & 0   & 19  & 0   & 64  & 0 \\
$\gamma_{12}$ & 0   & 9   & 3   & 79  & 0 \\
\hline%
\end{tabular}
\end{table}

\begin{table}[ht]\centering
\caption{\label{tbl:Ergebnisse:MOLSD:Dmax:VNS}Significance test
results for $D_{2}$ over all 100 test instances.}
\begin{tabular}{lrrrrr}\\%
\hline%
$\gamma$      & 1-BSH & 1-FSH & 1-EX & MOVNS/3        & MOVNS/9\\
\hline
$\gamma_{1}$  & 0   & 10  & 2   & 58  & 0 \\
$\gamma_{2}$  & 5   & 5   & 0   & 58  & 0 \\
$\gamma_{3}$  & 1   & 6   & 0   & 64  & 0 \\
$\gamma_{4}$  & 1   & 12  & 1   & 46  & 0 \\
$\gamma_{5}$  & 0   & 14  & 0   & 48  & 0 \\
$\gamma_{6}$  & 5   & 0   & 2   & 42  & 0 \\
$\gamma_{7}$  & 0   & 18  & 1   & 66  & 1 \\
$\gamma_{8}$  & 0   & 20  & 0   & 64  & 1 \\
$\gamma_{9}$  & 4   & 9   & 1   & 66  & 0 \\
$\gamma_{10}$ & 2   & 10  & 2   & 68  & 0 \\
$\gamma_{11}$ & 1   & 21  & 1   & 67  & 1 \\
$\gamma_{12}$ & 3   & 15  & 17  & 60  & 1 \\
\hline%
\end{tabular}
\end{table}

The configuration of MOVNS with all presented neighbourhood
definitions MOVNS/9 does not lead to satisfying results.
Integrating relatively weaker neighbourhood operators can
consequently not be regarded as a promising concept of hybridising
local search heuristics.

The running times for the different algorithms are almost
identical as the approaches differ only in terms of the additional
choice of the neighbourhood in MOVNS. On a Intel Pentium 4 PC
running at 1.8 GHz and 256 MB RAM, neighbouring solutions may be
computed within 0.0003 milliseconds, while the evaluation of a
solution takes 0.0402 milliseconds. The comparison of the
resolution behaviour of different neighbourhood operators shows no
significant difference in terms of running time.

\section{Conclusions}
A study of local search neighbourhoods for multi objective flow
shop scheduling has been presented. First experiments involving
twelve different combinations of optimality criteria revealed that
no single operator is superior for all considered problem
instance. Furthermore, it has been possible to show that the
identification probability of a specific Pareto optimal
alternative depends on the choice of the neighbourhood in the
local search procedure.

Improvements have been obtained in a significantly large
number of test cases by applying a concept of randomised variable
neighbourhood search. While the in average superior behaviour of
the MOVNS approach is independent from the choice of the
optimality criteria, a restriction to favourable neighbourhood
definitions is necessary as weaker operators do not contribute to
the quality of the obtained results.

The study demonstrates the advantages of hybridising traditional
local search techniques with multi neighbourhood strategies. This
is especially of importance as in the proposed setting of
alternatively selecting operators the computational complexity of
the approach is not increased while superior results are obtained.

\section*{Acknowledgements}
The author wants to thank two anonymous referees for their helpful
comments.

\bibliographystyle{plain}
\bibliography{lit_bank,lit_bank_nv,lit_bank_datei}

\end{document}